\documentclass[11pt]{article}
\usepackage[draft]{hyperref}
\usepackage{acl2016}
\usepackage{natbib}
\usepackage{times}
\usepackage{latexsym}
\usepackage{amsmath}
\usepackage{amssymb}
\usepackage{booktabs}
\usepackage{graphicx}
\usepackage{subcaption}
\usepackage{multirow}
\usepackage{algorithm}
\usepackage{algorithmic}
\usepackage{ifthen}
\usepackage{color}
\usepackage[utf8]{inputenc}
\usepackage{scabby-lite}
\usepackage{multicol}

\providecommand\sa{\ensuremath{\mathcal{a}}}
\providecommand\sd{\ensuremath{\mathcal{d}}}
\providecommand\se{\ensuremath{\mathcal{e}}}
\providecommand\sg{\ensuremath{\mathcal{g}}}
\providecommand\sh{\ensuremath{\mathcal{h}}}

\providecommand\sj{\ensuremath{\mathcal{j}}}
\providecommand\sk{\ensuremath{\mathcal{k}}}
\providecommand\sm{\ensuremath{\mathcal{m}}}
\providecommand\sn{\ensuremath{\mathcal{n}}}
\providecommand\so{\ensuremath{\mathcal{o}}}
\providecommand\sq{\ensuremath{\mathcal{q}}}
\providecommand\sr{\ensuremath{\mathcal{r}}}
\providecommand\st{\ensuremath{\mathcal{t}}}
\providecommand\su{\ensuremath{\mathcal{u}}}
\providecommand\sv{\ensuremath{\mathcal{v}}}
\providecommand\sw{\ensuremath{\mathcal{w}}}
\providecommand\sx{\ensuremath{\mathcal{x}}}
\providecommand\sy{\ensuremath{\mathcal{y}}}
\providecommand\sz{\ensuremath{\mathcal{z}}}
\providecommand\sA{\ensuremath{\mathcal{A}}}
\providecommand\sB{\ensuremath{\mathcal{B}}}
\providecommand\sC{\ensuremath{\mathcal{C}}}
\providecommand\sD{\ensuremath{\mathcal{D}}}
\providecommand\sE{\ensuremath{\mathcal{E}}}
\providecommand\sF{\ensuremath{\mathcal{F}}}
\providecommand\sG{\ensuremath{\mathcal{G}}}
\providecommand\sH{\ensuremath{\mathcal{H}}}
\providecommand\sI{\ensuremath{\mathcal{I}}}
\providecommand\sJ{\ensuremath{\mathcal{J}}}
\providecommand\sK{\ensuremath{\mathcal{K}}}
\providecommand\sL{\ensuremath{\mathcal{L}}}
\providecommand\sM{\ensuremath{\mathcal{M}}}
\providecommand\sN{\ensuremath{\mathcal{N}}}
\providecommand\sO{\ensuremath{\mathcal{O}}}
\providecommand\sP{\ensuremath{\mathcal{P}}}
\providecommand\sQ{\ensuremath{\mathcal{Q}}}
\providecommand\sR{\ensuremath{\mathcal{R}}}
\providecommand\sS{\ensuremath{\mathcal{S}}}
\providecommand\sT{\ensuremath{\mathcal{T}}}
\providecommand\sU{\ensuremath{\mathcal{U}}}
\providecommand\sV{\ensuremath{\mathcal{V}}}
\providecommand\sW{\ensuremath{\mathcal{W}}}
\providecommand\sX{\ensuremath{\mathcal{X}}}
\providecommand\sY{\ensuremath{\mathcal{Y}}}
\providecommand\sZ{\ensuremath{\mathcal{Z}}}
\providecommand\ba{\ensuremath{\mathbf{a}}}
\providecommand\bb{\ensuremath{\mathbf{b}}}
\providecommand\bc{\ensuremath{\mathbf{c}}}
\providecommand\bd{\ensuremath{\mathbf{d}}}
\providecommand\be{\ensuremath{\mathbf{e}}}
\providecommand\bg{\ensuremath{\mathbf{g}}}
\providecommand\bh{\ensuremath{\mathbf{h}}}
\providecommand\bi{\ensuremath{\mathbf{i}}}
\providecommand\bj{\ensuremath{\mathbf{j}}}
\providecommand\bk{\ensuremath{\mathbf{k}}}
\providecommand\bl{\ensuremath{\mathbf{l}}}
\providecommand\bn{\ensuremath{\mathbf{n}}}
\providecommand\bo{\ensuremath{\mathbf{o}}}
\providecommand\bp{\ensuremath{\mathbf{p}}}
\providecommand\bq{\ensuremath{\mathbf{q}}}
\providecommand\br{\ensuremath{\mathbf{r}}}
\providecommand\bs{\ensuremath{\mathbf{s}}}
\providecommand\bt{\ensuremath{\mathbf{t}}}
\providecommand\bu{\ensuremath{\mathbf{u}}}
\providecommand\bv{\ensuremath{\mathbf{v}}}
\providecommand\bw{\ensuremath{\mathbf{w}}}
\providecommand\bx{\ensuremath{\mathbf{x}}}
\providecommand\by{\ensuremath{\mathbf{y}}}
\providecommand\bz{\ensuremath{\mathbf{z}}}
\providecommand\bA{\ensuremath{\mathbf{A}}}
\providecommand\bB{\ensuremath{\mathbf{B}}}
\providecommand\bC{\ensuremath{\mathbf{C}}}
\providecommand\bD{\ensuremath{\mathbf{D}}}
\providecommand\bE{\ensuremath{\mathbf{E}}}
\providecommand\bF{\ensuremath{\mathbf{F}}}
\providecommand\bG{\ensuremath{\mathbf{G}}}
\providecommand\bH{\ensuremath{\mathbf{H}}}
\providecommand\bI{\ensuremath{\mathbf{I}}}
\providecommand\bJ{\ensuremath{\mathbf{J}}}
\providecommand\bK{\ensuremath{\mathbf{K}}}
\providecommand\bL{\ensuremath{\mathbf{L}}}
\providecommand\bM{\ensuremath{\mathbf{M}}}
\providecommand\bN{\ensuremath{\mathbf{N}}}
\providecommand\bO{\ensuremath{\mathbf{O}}}
\providecommand\bP{\ensuremath{\mathbf{P}}}
\providecommand\bQ{\ensuremath{\mathbf{Q}}}
\providecommand\bR{\ensuremath{\mathbf{R}}}
\providecommand\bS{\ensuremath{\mathbf{S}}}
\providecommand\bT{\ensuremath{\mathbf{T}}}
\providecommand\bU{\ensuremath{\mathbf{U}}}
\providecommand\bV{\ensuremath{\mathbf{V}}}
\providecommand\bW{\ensuremath{\mathbf{W}}}
\providecommand\bX{\ensuremath{\mathbf{X}}}
\providecommand\bY{\ensuremath{\mathbf{Y}}}
\providecommand\bZ{\ensuremath{\mathbf{Z}}}
\providecommand\Ba{\ensuremath{\mathbb{a}}}
\providecommand\Bb{\ensuremath{\mathbb{b}}}
\providecommand\Bc{\ensuremath{\mathbb{c}}}
\providecommand\Bd{\ensuremath{\mathbb{d}}}
\providecommand\Be{\ensuremath{\mathbb{e}}}
\providecommand\Bf{\ensuremath{\mathbb{f}}}
\providecommand\Bg{\ensuremath{\mathbb{g}}}
\providecommand\Bh{\ensuremath{\mathbb{h}}}
\providecommand\Bi{\ensuremath{\mathbb{i}}}
\providecommand\Bj{\ensuremath{\mathbb{j}}}
\providecommand\Bk{\ensuremath{\mathbb{k}}}
\providecommand\Bl{\ensuremath{\mathbb{l}}}
\providecommand\Bm{\ensuremath{\mathbb{m}}}
\providecommand\Bn{\ensuremath{\mathbb{n}}}
\providecommand\Bo{\ensuremath{\mathbb{o}}}
\providecommand\Bp{\ensuremath{\mathbb{p}}}
\providecommand\Bq{\ensuremath{\mathbb{q}}}
\providecommand\Br{\ensuremath{\mathbb{r}}}
\providecommand\Bs{\ensuremath{\mathbb{s}}}
\providecommand\Bt{\ensuremath{\mathbb{t}}}
\providecommand\Bu{\ensuremath{\mathbb{u}}}
\providecommand\Bv{\ensuremath{\mathbb{v}}}
\providecommand\Bw{\ensuremath{\mathbb{w}}}
\providecommand\Bx{\ensuremath{\mathbb{x}}}
\providecommand\By{\ensuremath{\mathbb{y}}}
\providecommand\Bz{\ensuremath{\mathbb{z}}}
\providecommand\BA{\ensuremath{\mathbb{A}}}
\providecommand\BB{\ensuremath{\mathbb{B}}}
\providecommand\BC{\ensuremath{\mathbb{C}}}
\providecommand\BD{\ensuremath{\mathbb{D}}}
\providecommand\BE{\ensuremath{\mathbb{E}}}
\providecommand\BF{\ensuremath{\mathbb{F}}}
\providecommand\BG{\ensuremath{\mathbb{G}}}
\providecommand\BH{\ensuremath{\mathbb{H}}}
\providecommand\BI{\ensuremath{\mathbb{I}}}
\providecommand\BJ{\ensuremath{\mathbb{J}}}
\providecommand\BK{\ensuremath{\mathbb{K}}}
\providecommand\BL{\ensuremath{\mathbb{L}}}
\providecommand\BM{\ensuremath{\mathbb{M}}}
\providecommand\BN{\ensuremath{\mathbb{N}}}
\providecommand\BO{\ensuremath{\mathbb{O}}}
\providecommand\BP{\ensuremath{\mathbb{P}}}
\providecommand\BQ{\ensuremath{\mathbb{Q}}}
\providecommand\BR{\ensuremath{\mathbb{R}}}
\providecommand\BS{\ensuremath{\mathbb{S}}}
\providecommand\BT{\ensuremath{\mathbb{T}}}
\providecommand\BU{\ensuremath{\mathbb{U}}}
\providecommand\BV{\ensuremath{\mathbb{V}}}
\providecommand\BW{\ensuremath{\mathbb{W}}}
\providecommand\BX{\ensuremath{\mathbb{X}}}
\providecommand\BY{\ensuremath{\mathbb{Y}}}
\providecommand\BZ{\ensuremath{\mathbb{Z}}}
\providecommand\balpha{\ensuremath{\mbox{\boldmath$\alpha$}}}
\providecommand\bbeta{\ensuremath{\mbox{\boldmath$\beta$}}}
\providecommand\btheta{\ensuremath{\mbox{\boldmath$\theta$}}}
\providecommand\bphi{\ensuremath{\mbox{\boldmath$\phi$}}}
\providecommand\bpi{\ensuremath{\mbox{\boldmath$\pi$}}}
\providecommand\bpsi{\ensuremath{\mbox{\boldmath$\psi$}}}
\providecommand\bmu{\ensuremath{\mbox{\boldmath$\mu$}}}

\DeclareMathOperator*{\sign}{sign}
\DeclareMathOperator*{\diag}{diag} 
\newcommand\inv[1]{\ensuremath{\frac{1}{#1}}}
\newcommand\half{\ensuremath{\frac{1}{2}}}

\newcommand\refsec[1]{Section~\ref{sec:#1}}

\newcommand\reffig[1]{Figure~\ref{fig:#1}}

\newcommand\reftab[1]{Table~\ref{tab:#1}}

\ifthenelse{\isundefined{\definition}}{\newtheorem{definition}{Definition}}{}
\ifthenelse{\isundefined{\assumption}}{\newtheorem{assumption}{Assumption}}{}
\ifthenelse{\isundefined{\hypothesis}}{\newtheorem{hypothesis}{Hypothesis}}{}
\ifthenelse{\isundefined{\proposition}}{\newtheorem{proposition}{Proposition}}{}
\ifthenelse{\isundefined{\theorem}}{\newtheorem{theorem}{Theorem}}{}
\ifthenelse{\isundefined{\lemma}}{\newtheorem{lemma}{Lemma}}{}
\ifthenelse{\isundefined{\corollary}}{\newtheorem{corollary}{Corollary}}{}
\ifthenelse{\isundefined{\alg}}{\newtheorem{alg}{Algorithm}}{}
\ifthenelse{\isundefined{\example}}{\newtheorem{example}{Example}}{}
\newcommand{\E}{\ensuremath{\mathbb{E}}} 
\newcommand\KL[2]{\ensuremath{\text{KL}\left( #1 \| #2 \right)}} 

\newcommand{\ttuples}{\mathrm{tuples}}
\newcommand{\tvalue}{\mathrm{value}}
\newcommand{\tunit}{\mathrm{unit}}
\newcommand{\tdescription}{\mathrm{description}}

\newcommand{\wvec}{\operatorname{wvec}}
\newcommand{\kb}{\mathcal{K}}

\newcommand{\fone}{F$_1$}

\newcommand{\systemb}{\textsc{baseline}}

\newcommand{\systemo}{\textsc{lr+rnn}}

\aclfinalcopy 
\def\aclpaperid{706} 

\title{How Much is 131 Million Dollars? Putting Numbers in Perspective with Compositional Descriptions}

\author{%
  Arun Tejasvi Chaganty \\
  Computer Science Department \\
  Stanford University \\
  {\tt chaganty@cs.stanford.edu}
\And%
	Percy Liang \\
  Computer Science Department \\
  Stanford University \\
  {\tt pliang@cs.stanford.edu}
}

\date{}

\begin{document}

\maketitle

\begin{abstract}
How much is 131 million US dollars?  
To help readers put such numbers in context, we propose a new task of automatically generating short descriptions known as perspectives, 
e.g.\ ``\$131 million is about the cost to employ everyone in Texas over a lunch period''.
First, we collect a dataset of numeric mentions in news articles,
where each mention is labeled with a set of rated perspectives.
We then propose a system to generate these descriptions consisting of two steps: formula construction and description generation.  
In construction, we compose formulae from numeric facts in a knowledge base and rank the resulting formulas based on familiarity, numeric proximity and semantic compatibility. 
In generation, we convert a formula into natural language using a sequence-to-sequence recurrent neural network.
Our system obtains a $15.2\%$ \fone{} improvement over a non-compositional baseline at formula construction and 
a 12.5 BLEU point improvement over a baseline description generation.


\end{abstract}

\section{Introduction}

\begin{figure}[ht] 
  \begin{center} 
    \includegraphics[scale=1]{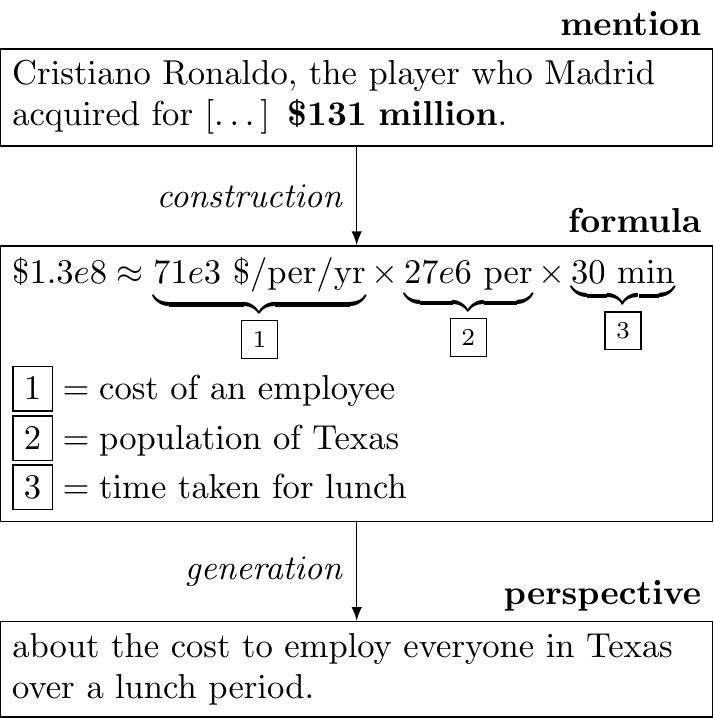} 
  \end{center} 
  \caption{\label{fig:overview} An overview of the perspective generation task: 
given a \textit{numeric mention}, generate a short description (a \textit{perspective}) that allows the reader to appreciate the scale of the mentioned number. 
In our system, we first construct a \textit{formula} over facts in our knowledge base and then generate a description of that formula.
}
\end{figure}

When posed with a mention of a number, such as ``Cristiano Ronaldo, the player who Madrid acquired for [\dots] a \textit{\$131 million}'' (\reffig{overview}), it is often difficult to comprehend the scale of large (or small) absolute values like \textit{\$131 million} \citep{paulos1988innumeracy,seife2010proofiness}. 
Studies have shown that providing relative comparisons, or \emph{perspectives}, 
such as ``about the cost to employ everyone in Texas over a lunch period''
significantly improves comprehension when measured in terms of memory retention or outlier detection~\citep{barrio2016comprehension}.

Previous work in the HCI community 
has relied on either manually generated perspectives~\citep{barrio2016comprehension} or 
present a fact as is from a knowledge base~\citep{chiachieri2013dictionary}.
As a result, these approaches are limited to contexts in which a relevant perspective already exists. 


In this paper, we generate perspectives by composing facts from a knowledge base.
For example, we might describe \textit{\$100,000} to be ``about twice \textul{the median income} for \textul{a year}'', and describe \textit{\$5 million} to be the ``about \textul{how much the average person makes} over \textul{their lifetime}''.
Leveraging compositionality allows us to achieve broad coverage of numbers from a relatively small collection of familiar facts, e.g.\ median income and a person's lifetime.

Using compositionality in perspectives is also concordant with our understanding of how people learn to appreciate scale. \citet{jones2009scale} find that students learning to appreciate scale do so mainly by 
\textit{anchoring} with familiar concepts, e.g.\ \$50,000 is slightly less than the median income in the US, 
and by \textit{unitization}, i.e.\ improvising a system of units that is more relatable, e.g.\ using the Earth as a measure of mass when describing the mass of Jupiter to be that of 97 Earths.
Here, compositionality naturally unitizes the constituent facts: in the examples above, money was unitized in terms of median income, and time was unitized in a person's lifetime.
Unitization and anchoring have also been proposed by \citet{chevalier2013composition} as the basis of a design methodology for constructing visual perspectives called concrete scales.

When generating compositional perspectives, we must address two key challenges: constructing familiar, relevant and meaningful formulas and generating easy-to-understand descriptions or perspectives.
We tackle the first challenge using an overgenerate-and-rank paradigm, selecting formulas using signals from familiarity, compositionality, numeric proximity and semantic similarity.
We treat the second problem of generation as a translation problem and use a sequence-to-sequence recurrent neural network (RNN) to generate perspectives from a formula.

We evaluate individual components of our system quantitatively on a dataset collected using crowdsourcing. 
Our formula construction method improves on \fone{} over
a non-compositional baseline by about 17.8\%.
Our generation method improves over a simple baseline by 12.5 BLEU points.

\section{Problem statement}

The input to the \emph{perspective generation} task is a sentence $s$ containing a \textit{numeric mention} $x$: a span of tokens within the sentence which describes a quantity with value $x.\tvalue$ and of unit $x.\tunit$.
In \reffig{overview}, the numeric mention $x$ is ``\$131 million'', $x.\tvalue
= 1.31e8$ and $x.\tunit = \text{\$}$.
The output is a description $y$ that puts $x$ in perspective.

We have access to a knowledge base $\kb$ with numeric tuples $t = (t.\tvalue, t.\tunit, t.\tdescription)$.
\reftab{kb} has a few examples of tuples in our knowledge base.
Units (e.g. $\text{\$/per/yr}$) are fractions composed either of fundamental
units (length, area, volume, mass, time) or of ordinal units (e.g.\ cars,
people, etc.). 

\begin{table}
  \begin{tabular}{p{.64\columnwidth} p{.30\columnwidth}}
    \toprule
\bf Description & \bf Value Unit \\ \midrule
cost of an employee                              &      71e3 \$/year/person \\
population of Texas                              &      27e3 person \\
number of employees at Google                    &      57e3 person \\
average household size                           &      2.54 person \\
time taken for a basketball game                 &        60 minute \\
average lifetime for a person                    &        79 year \\
a week                                           &         1 week \\
time taken for lunch                             &        30 minute \\
cost of property in the Bay area                 &       1e3 \$/ft$^2$ \\
area of a city block                             &      10e3 $m^2$ \\
\bottomrule
  \end{tabular}

  \caption{\label{tab:kb} A subset of our knowledge base of numeric tuples.
  Tuples with fractional units (e.g.\ \$/ft$^2$) can be combined with other tuples to create formulas.
  }
\end{table}

The first step of our task, described in \refsec{selection},
is to construct a \textit{formula} $f$ over numeric tuples in $\kb$
that has the same value and unit as the numeric mention $x$.
A valid formula comprises of an arbitrary multiplier $f.m$ and a sequence of tuples $f.\ttuples$.
The value of a formula, $f.\tvalue$, is simply the product of the multiplier and the values of the tuples,
and the unit of the formula, $f.\tunit$, is the product of the units of the tuples.
In \reffig{overview}, the formula has a multiplier of $1$ and is composed of tuples $\fbox{1}$, $\fbox{2}$ and $\fbox{3}$;
it has a value of $1.3e8$ and a unit of $\text{\$}$.

The second step of our task, described in \refsec{generation}, is to generate a
\textit{perspective} $y$, a short noun phrase that realizes $f$.
Typically, the utterance will be formed using variations of the descriptions of the tuples in $f.\ttuples$.






\section{Dataset construction}
\label{sec:dataset}

We break our data collection task into two steps, mirroring formula selection
and description generation: first, we collect descriptions of formulas
constructed exhaustively from our knowledge base (for generation), and then we use these
descriptions to collect preferences for perspectives (for construction).


\paragraph{Collecting the knowledge base.}
We manually constructed a knowledge base with 142 tuples 
and 9 fundamental units%
\footnote{Namely, length, area, volume, time, weight, money, people, cars and guns. These units were chosen because they were well represented in the corpus.}
from 
the United States Bureau of Statistics,
the orders of magnitude topic on Wikipedia
and other Wikipedia pages.
The facts chosen are somewhat crude; for example, though ``the cost of an employee'' is a very context dependent quantity, we take its value to be the median cost for an employer in the United States, \$71,000.
Presenting facts at a coarse level of granularity makes them more familiar to the general reader while still being appropriate for perspective generation: the intention is to convey the right scale, not necessarily the precise quantity. 

\begin{figure}[t]
  \begin{center}
    \includegraphics[width=\columnwidth]{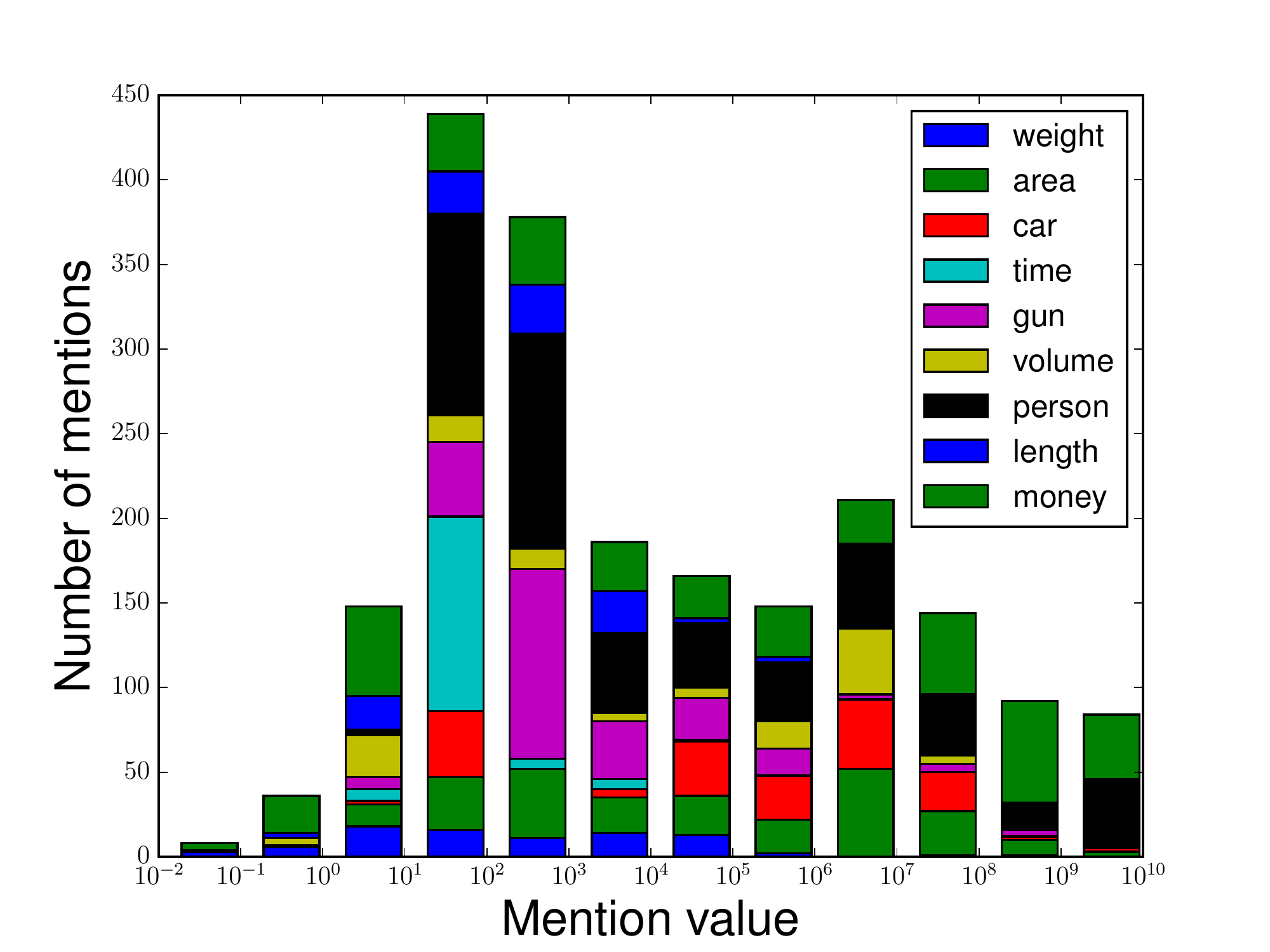}
  \end{center}
  \caption{\label{fig:mentions-histogram} A histogram of the absolute values of numeric mentions by type. There are 100--300 mentions of each unit.}
\end{figure}

\paragraph{Collecting numeric mentions.}
We collected 53,946 sentences containing numeric mentions from the newswire section of LDC2011T07
using simple regular expression patterns like \texttt{\$([0-9]+(,[0-9]+)*(.[0-9]+)? ((hundred)|(thousand)|(million)| (billion)|(trillion)))}.
The values and units of the numeric mentions in each sentence were normalized
and converted to fundamental units (e.g.\ from miles to length). 
We then randomly selected up to 200 mentions of each of the 9 types in bins with boundaries ${10}^{-3}, 1, {10}^3, {10}^6, {10}^9, {10}^{12}$ leading to 4,931 mentions that are stratified by unit and magnitude.\footnote{Some types had fewer than 200 mentions for some bins.} 
Finally, we chose mentions which could be described by at least one numeric expression, resulting in the 2,041 mentions that we use in our experiments (\reffig{mentions-histogram}).  
We note that there is a slight bias towards mentions of money and people because these are more common in the news corpus.

\paragraph{Generating formulas.}

\begin{figure}[t]
  \begin{center}
    \includegraphics[width=\columnwidth]{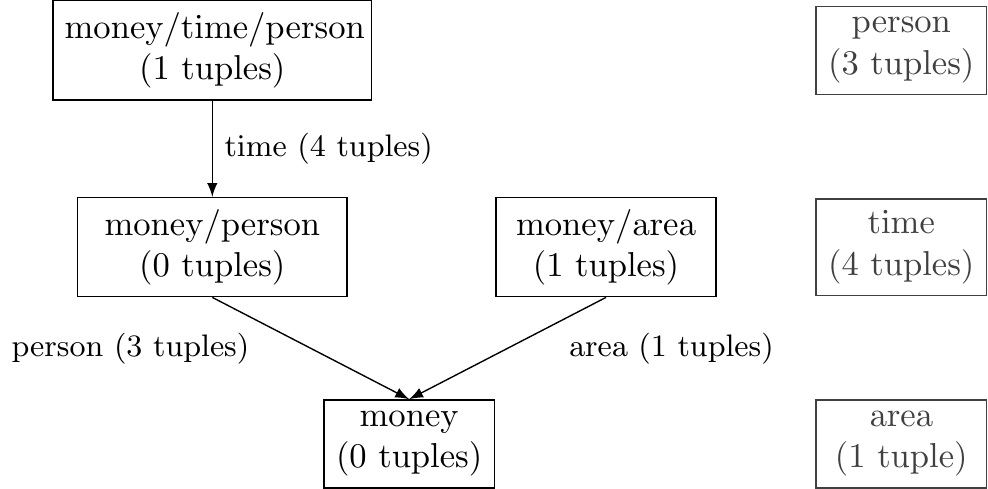}
  \end{center}
  \caption{\label{fig:unit-graph} The graph over tuples generated from the knowledge base subset in \reftab{kb}.}
\end{figure}

Next, we exhaustively generate valid formulas from our knowledge base.
We represent the knowledge base as a graph over units with vertices and edges
annotated with tuples (\reffig{unit-graph}).
Every vertex in this graph is labeled with a unit $u$ and contains the set of tuples with this unit: $\{t \in \kb: t.\tunit = u\}$.
Additionally, for every vertex in the graph with a unit of the form $u_1 / u_2$,
where $u_2$ has no denominator, we add an edge from $u_1/u_2$ to $u_1$,
annotated with all tuples of type $u_2$:
in \reffig{unit-graph} we add an edge from \textit{money/person} to \textit{money}
annotated with the three person tuples in \reftab{kb}.
The set of formulas with unit $u$ is obtained by enumerating all paths in the
graph which terminate at the vertex $u$.
The multiplier of the formula is set so that the value of the formula matches the value of the mention.
For example, the formula in \reffig{overview} was constructed by traversing the graph from \textit{money/time/person} to \textit{money}: we start with a tuple in  \textit{money/time/person} (\textit{cost of an employee}) and then multiply by a tuple with unit \textit{time} (\textit{time for lunch}) and then by unit \textit{person} (\textit{population of Texas}),
thus traversing two edges to arrive at \textit{money}.

Using the 142 tuples in our knowledge base, we generate a total of 1,124 formulas sans multiplier.

\paragraph{Collecting descriptions of formulas.}

\begin{figure}[t]
  \begin{center}
    \includegraphics[width=\columnwidth]{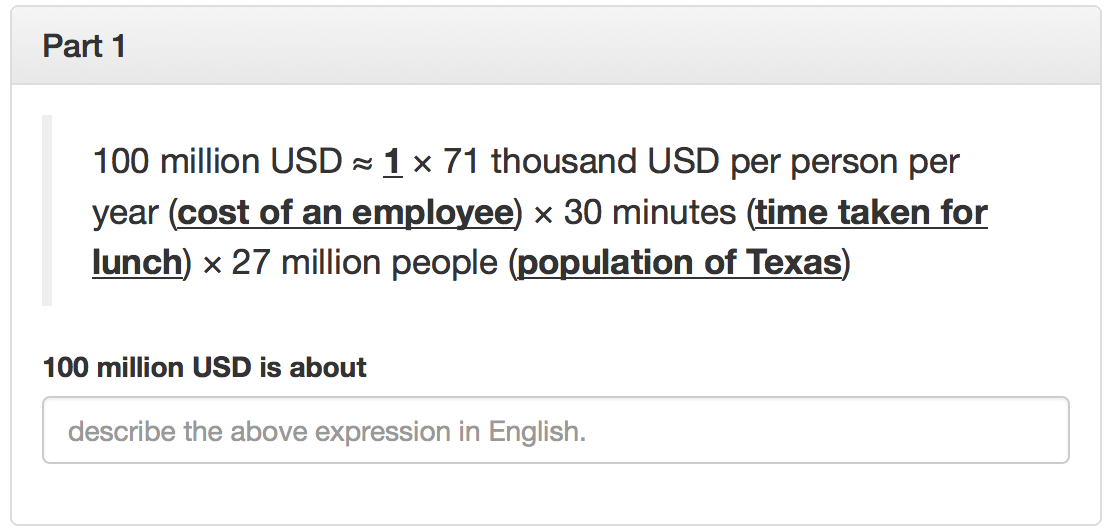}
  \end{center}
  \caption{\label{fig:expr-task} 
  A screenshot of the crowdsourced task to generate natural language descriptions, or perspectives, from formulas.}
\end{figure}

The main goal of collecting descriptions of formulas is to train a language generation system, though these descriptions will also be useful while collecting training data for formula selection.
For every unit in our knowledge base and every value in the set $\{10^{-7},
10^{-6} \ldots, 10^{10}\}$, we generated all valid formulas.
We further restricted this set to formulas with a multiplier between $1/100$ and $100$, based on the rationale that human cognition of scale sharply drops beyond an order of magnitude~\citep{tretter2006accuracy}.
In total, 5000 formulas were presented to crowdworkers on Amazon Mechanical Turk, with a prompt asking them to rephrase the formula as an English expression (\reffig{expr-task}).%
\footnote{Crowdworkers were paid \$0.08 per description.}
We obtained 5--7 descriptions of each formula, leading to a total of 31,244 unique descriptions.

%

\paragraph{Collecting data on formula preference.}

\begin{figure}[t]
  \begin{center}
    \includegraphics[width=\columnwidth]{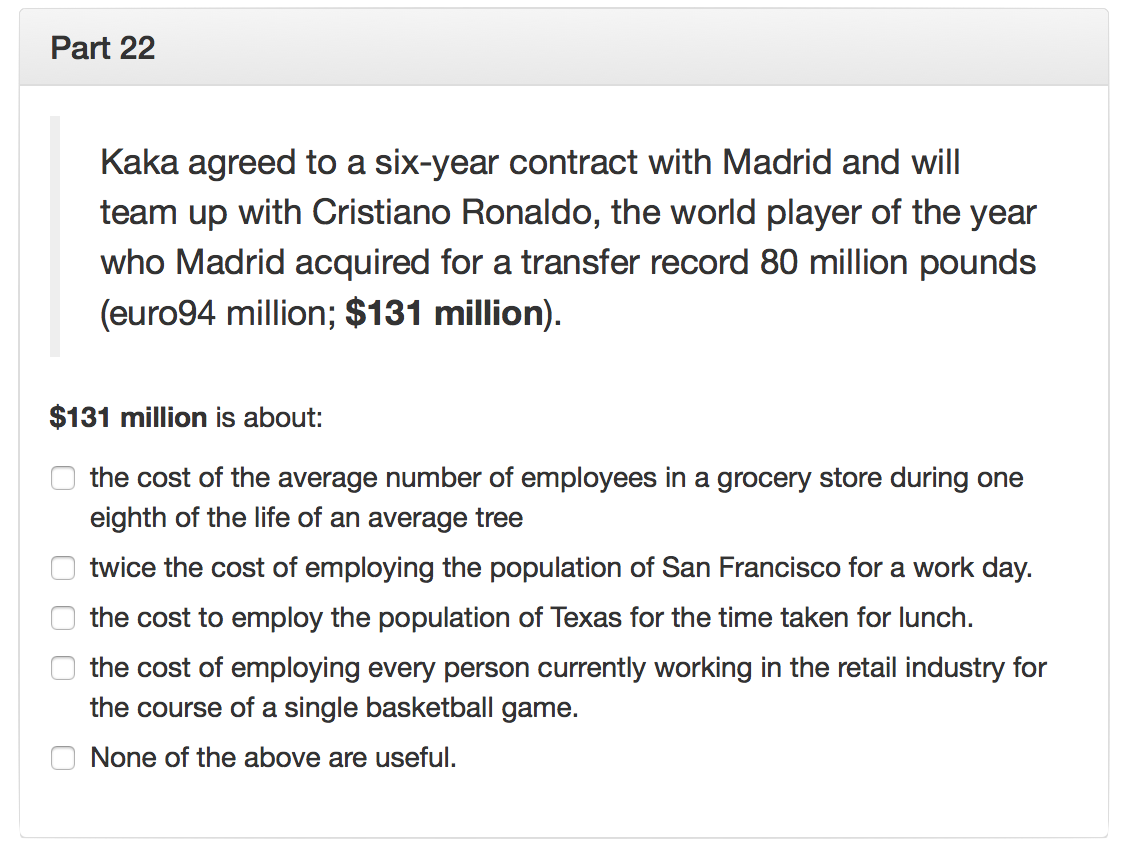}
  \end{center}
  \caption{\label{fig:rank-task} 
  A screenshot of the crowdsourced task to identify which formulas are useful to crowdworkers in understanding the highlighted mentioned number.}
\end{figure}

Finally, given a numeric mention, we ask crowdworkers which perspectives from the description dataset they prefer.
Note that formulas generated for a particular mention may differ in multiplier with a formula in the description dataset.
We thus relax our constraints on factual accuracy while collecting this formula preference dataset:
for each mention $x$, 
we choose a random perspective from the description dataset described above
corresponding to a formula whose value is within a factor of 2 
from the
mention's value, $x.\tvalue$. 
A smaller factor led to too many mentions without a valid comparison, while a larger one led to blatant factual inaccuracies.
The perspectives were partitioned into sets of four 
and displayed to crowdworkers
along with a ``None of the above'' option 
with the following prompt: ``We would like you to pick up to two of these descriptions that are useful in understanding the scale of the highlighted number''
(\reffig{rank-task}).
A formula is rated to be useful by simple majority.\footnote{Crowdworkers were paid \$0.06 to vote on each set of perspectives.}

\begin{figure}[t]
  \begin{center}
    \includegraphics[width=\columnwidth]{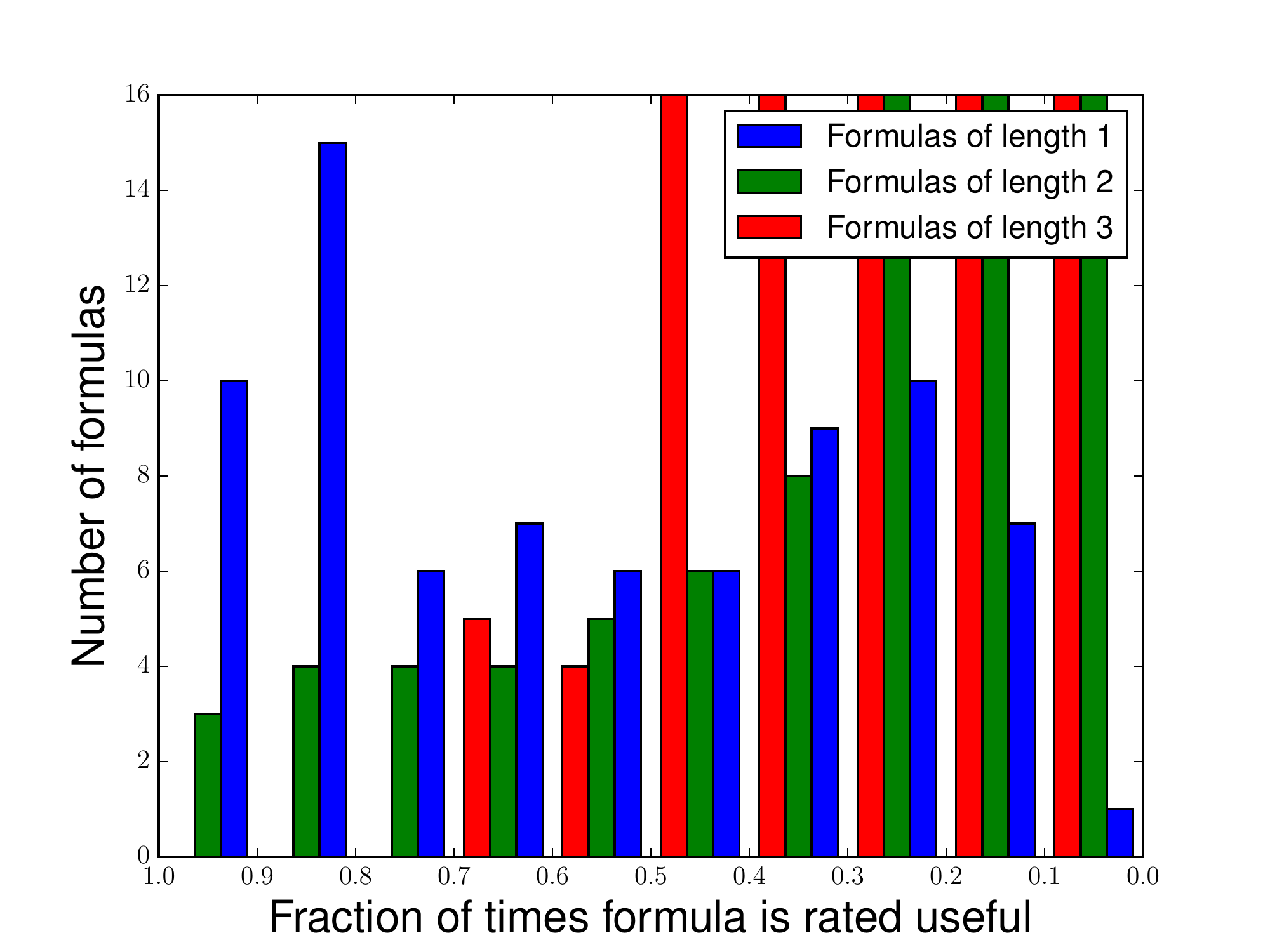}
  \end{center}
  \caption{\label{fig:fertility-expr} 
  A histogram comparing formula length to ratings of usefulness (clipped for readability). 
  Non-compositional perspectives with a single tuple are broadly useful.
  Useful compositional perspectives tend to be more context-specific than non-compositional ones, and 
  many of the formulas that can be generated from the knowledge base are spurious.
  }
\end{figure}

\reffig{fertility-expr} provides a summary of the dataset collected, visualizing how many formulas are useful, controlling for the size of the formula.
The exhaustive generation procedure produces a large number of spurious formulas like ``20 $\times$ trash generated in the US $\times$ a minute $\times$ number of employees on Medicare''.
Nonetheless, compositional formulas are quite useful in the appropriate context; \reftab{examples} presents some mentions with highly rated perspectives and formulas.


\begin{table*}[t]
  \begin{center}
\begin{tabular}{p{0.34\textwidth} p{0.29\textwidth} p{0.29\textwidth}}
  \toprule
 \bf  Sentence &\bf  That's about \dots &\bf  Formula \\
  \midrule
  The Billings-based Stillwater Mining produced \textbf{601,000 ounces of platinum}. &
  4 times the weight of an elephant. &
  4 $\times$ weight of an elephant.
  \\
Authorities estimate there are about \textbf{60 million guns} in Yemen. & twice the gun ownership of the population of Texas   &
2 $\times$ gun ownership $\times$ population of Texas 
\\ 
Water is flowing into Taihu lake at a rate of \textbf{150 cubic meters} per second. &
  how much water would flow from a tap left on for a week. &
rate of flow of water from tap  $\times$ a week
  \\
The bank had held auctions, selling around \textbf{US\$1 billion} worth of three-month bills.
& half the cost of employing the population of Texas for a work day. &
1/2 $\times$ cost of an employee $\times$ time taken for a work day $\times$ population of Texas
\\
The government[s] have promised to rent about \textbf{1.2 million sq. feet}. &
the area of forest logged in a single minute & 
90 $\times$ area of forest logged $\times$ a minute
\\
  \bottomrule
\end{tabular}

  \end{center}
  \caption{\label{tab:examples} Examples of numeric mentions, perspectives and their corresponding formulas in the dataset.
  All the examples except the last one are rated to be useful by crowdworkers.
  }
\end{table*}

\section{Formula selection}
\label{sec:selection}

We now turn to the first half of our task: given a numeric mention $x$ and a knowledge base $\kb$,
select a formula $f$ over $\kb$ with the same value and unit as the mention.
It is easy to generate a very large number of formulas for any mention.
For the example, ``Cristiano Ronaldo, the player who Madrid acquired for [\dots] \$131 million.'',
the small knowledge base in \reftab{kb} can generate the 12 different formulas,%
\footnote{The full knowledge base described in \refsec{dataset} can generate 242 formulas with the unit money (sans multiplier).}
including the following:

\begin{enumerate}
  \setlength\itemsep{0pt}
  \item \small $1$ $\times$ the cost of an employee $\times$ the population of Texas $\times$ the time taken for lunch.
  \item \small $400$ $\times$ the cost of an employee $\times$ average household size $\times$ a week.
  \item \small $1$ $\times$ the cost of an employee $\times$ number of employees at Google $\times$ a week.
  \item \small $1$ $\times$ cost of property in the Bay Area $\times$ area of a city block.
\end{enumerate}

Some of the formulas above are clearly worse than others:
the key challenge is picking a formula that will lead to a meaningful and relevant perspective.


\paragraph{Criteria for ranking formulas.}

\begin{table}
  \begin{center}
    \begin{tabular}{p{0.24\columnwidth} p{0.49\columnwidth}{l}}
  \toprule
  \bf Type &\bf  Features &\bf  \#\\ \midrule
  Proximity & \small $\sign(\log(f.m))$, $|\log(f.m)|$ & 1 \\
  Familiarity & \small $\BI[t]$  & 142 \\
  Compatibility &\small  $\BI[t,t']$ & 20022 \\
  Similarity &\small  ${\wvec(s)}^\top$ $\wvec(t.\tdescription)$ & 1 \\
  \bottomrule
\end{tabular}
  \end{center}
\caption{\label{tab:features} Feature templates used to score a formulas $f$ and their counts (\#), where 
$f.m$ is the formula's multiplier and $t, t' \in f.\ttuples$ are tuples in the formula.
}
\end{table}

We posit the following principles to guide our choice in features (\reftab{features}).

\textbf{Proximity:} \textit{A numeric perspective should be within an order of magnitude of the mentioned value}.
Conception of scale quickly fails with quantities that exceed ``human scales''~\citep{tretter2006accuracy}: numbers that are significantly away from $1/10$ and $10$.
We use this principle to prune formulas with multipliers not in the range $[1/100, 100]$ (e.g.\ example 2 above) and introduce features for numeric proximity.

\textbf{Familiarity:} 
\textit{A numeric perspective should be composed of concepts familiar to the reader}.
The most common technique cited by those who do well at scale cognition tests is reasoning in terms of familiar objects~\citep{tretter2006accuracy,jones2009scale,chevalier2013composition}.
Intuitively, the average American reader may not know exactly how many people are in Texas,
but is familiar enough with the quantity to effectively reason using Texas' population as a unit.
On the other hand, it is less likely that the same reader is familiar with even the concept of Angola's population. 

Of course, because it is so personal, familiarity is difficult to capture.
With additional information about the reader, e.g.\ their location,
it is possible to personalize the chosen tuples \citep{kim2016analogies}.
Without this information, we back off to a global preference on tuples by using indicator features for each tuple in the formula.

\textbf{Compatibility:} 
Similarly, some tuple combinations are more natural (``median income $\times$ a month'') while others are less so
(``weight of a person $\times$ population of Texas'').
We model compatibility between tuples in a formula using an indicator feature. 

\textbf{Similarity:} 
\textit{A numeric perspective should be relevant to the context}. 
Apart from helping with scale cognition, a perspective should also place the mentioned quantity in appropriate context:
for example, NASA's budget of \$17 billion 
could be described as 0.1\% of the United States' budget
or 
the amount of money it could cost to feed Los Angeles for a year. 
While both perspectives are appropriate, the former is more relevant than the latter.

We model context relevance using word vector similarity between the tuples of the formula and the sentence containing the mention as a proxy for semantic similarity.
Word vectors for a sentence or tuple description are computed by taking the mean of the word vectors for every non-stop-word token. The word vectors at the token level are computed using word2vec~\citep{mikolov2013efficient}.

\begin{table}
  \begin{center}
    \begin{tabular}{p{0.8\columnwidth} p{0.1\columnwidth}}
    \toprule
    \bf Formula & \bf Score \\ \midrule
    \multicolumn{2}{p{0.9\columnwidth}}{%
    Studies estimate \textbf{36,000 people} die on average each year from seasonal flu.
    } \\
 \midrule
 1/4 $\times$ global death rate $\times$ a day & 0.67 \\
 5 $\times$ death rate in the US $\times$ a day & 0.64 \\
 1/3 $\times$ number of employees at Microsoft & 0.60 \\

 \midrule
    \multicolumn{2}{p{0.9\columnwidth}}{%
    Gazprom's exports to Europe [\dots] will total \textbf{60 billion cubic meters} \dots
    } \\
 \midrule
    oil produced by the US $\times$ average lifetime & 0.78 \\
    average coffee consumption $\times$ population of the world $\times$ average lifetime & 0.78 \\
    2 $\times$ average coffee consumption $\times$ population of Asia $\times$ average lifetime & 0.73 \\
    \bottomrule
  \end{tabular}
  \end{center}
  \caption{\label{tab:examples-selection} The top three examples outputted by the ranking system with the scores reported by the system.
  }
\end{table}

\paragraph{Evaluation.}

We train a logistic regression classifier using the features described in \reftab{features}
using the perspective ratings collected in \refsec{dataset}.
Recall that the formula for each perspective in the dataset is assigned a positive (``useful'') label if it was labeled to be useful
to the majority of the workers.
\reftab{results-selection} presents results on classifying formulas as useful with a feature ablation.\footnote{%
Significance results are computed by the bootstrap test as described in \citet{kirkpatrick2012significance} using the output of classifiers trained on the entire training set.}  

Familiarity and compatibility are the most useful features when selecting formulas, each having a significant increase in \fone{} over the proximity baseline.
There are minor gains from combining these two features.
On the other hand, semantic similarity does not affect performance relative to the baseline.
We find that this is mainly due to the disproportionate number of unfamiliar formulas present in the dataset that drown out any signal.
\reftab{examples-selection} presents two examples of the system's ranking of formulas.

%


\begin{table*}
  \begin{center}
    \begin{subfigure}[t]{0.57\textwidth}
\begin{tabular}{l lll | lll}
  \toprule
  \bf Feature set & \multicolumn{3}{c}{\bf Train} & \multicolumn{3}{c}{\bf Dev} \\ 
  & P & R & \fone{} & P & R & \fone{} \\ \midrule
  Proximity     & 56.4 & 48.7 & 52.2 & 56.3 & 48.8 & 52.3 \\
  Similarity    & 65.1 & 34.9 & 45.4 & 65.1 & 34.9 & 45.4 \\ 
  Familiarity$^*$   & 70.5 & 63.5 & 66.8 & 69.6 & 62.9 & 66.1 \\
  Compatibility$^+$ & 66.9 & \textbf{74.4} & 70.4 & 65.4 & \textbf{73.1} & 69.0 \\
  F + C$^\dagger$         & \textbf{73.8} & 70.3 & \textbf{72.1} & \textbf{71.5} & 68.9 & \textbf{70.1} \\
  F + C + P$^\dagger$     & \textbf{73.8} & 70.3 & \textbf{72.1} & \textbf{71.5} & 68.9 & \textbf{70.1} \\
  F + C + P + S$^\dagger$ & \textbf{73.8} & 70.3 & 72.0 & 71.4 & 68.6 & 69.9 \\
  \bottomrule
\end{tabular}

        \caption{\label{tab:results-selection} the formula construction system.
        Precision, Recall and \fone{} are cross-validated on 10-folds. 
        $^*$significant \fone{} versus P and S with $p < 0.01$.
        $^+$significant \fone{} versus P, S and F with $p < 0.01$.
        $^\dagger$significant \fone{} versus P, S, F and C with $p < 0.05$.
        }
    \end{subfigure}%
    \hfill
    \begin{subfigure}[t]{0.36\textwidth}
\begin{tabular}{p{0.44\textwidth}p{0.19\textwidth}p{0.19\textwidth}}
  \toprule
  \bf System   & \bf Train BLEU & \bf Test BLEU \\ \midrule
  Baseline & 65.00      & 57.32 \\
  RNN$^*$ & \textbf{81.50} & \textbf{69.79} \\
  \bottomrule
\end{tabular}

        \caption{\label{tab:results-generation} the description generation system.
        $^*$significant BLEU score versus the baseline with $p < 0.01$.
        }
    \end{subfigure}%
  \end{center}
  \caption{\label{tab:results} Evaluation of perspective generation subsystems.
    }
\end{table*}

\section{Perspective generation}
\label{sec:generation}

Our next goal is to generate natural language descriptions, also known as perspectives, given a formula.
Our approach models the task as a sequence-to-sequence translation task from formulas to natural language.
We first describe a rule-based baseline
and then describe a recurrent neural network (RNN) with an attention-based copying mechanism~\citep{wang2016recombination}.

\paragraph{Baseline.}
As a simple approach to generate perspectives, we just combine tuples in the formula with the neutral prepositions
\textit{of} and \textit{for}, e.g.\ ``1/5th \textit{of} the cost of an employee \textit{for} the population of Texas \textit{for}
the time taken for lunch.''


\paragraph{Sequence-to-sequence RNN.}

\begin{figure}[t]
  \begin{center}
    \includegraphics[width=\columnwidth]{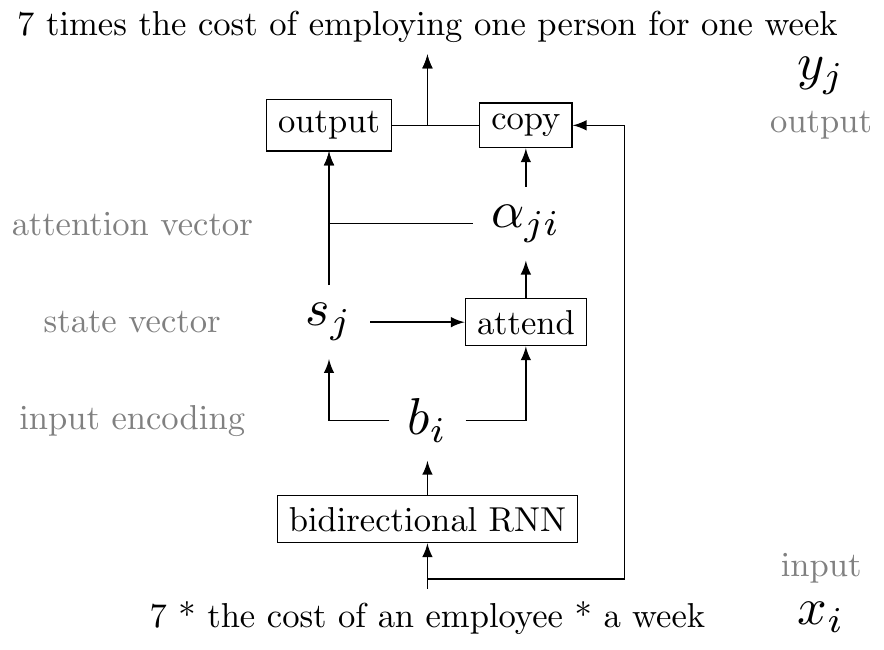}
  \end{center}
  \caption{\label{fig:seq-to-seq} We model description generation as a sequence transduction task,
    with input as formulas (at bottom) and output as
    perspectives (at top).
    We use a RNN with an attention-based copying mechanism.
  }
\end{figure}

We use formula-perspective pairs from the dataset to create a sequence-to-sequence task:
the input is composed using the formula's multiplier and descriptions of its
tuples connected with the symbol `*'; the output is the perspective (\reffig{seq-to-seq}).

Our system is based on the model described in \citet{wang2016recombination}.
Given a sequence of input tokens ($\bx = (x_i)$), the model computes a context-dependent vector ($\bb = (b_i)$) for each token using a bidirectional RNN with LSTM units.
We then generate the output sequence ($y_j$) left to right as follows.
At each output position, we have a hidden state vector ($s_j$)
which is used to produce an ``attention'' distribution ($\balpha_{j} = (\alpha_{ji})$) over input tokens: $\alpha_{ji} = \operatorname{Attend}(s_j, b_i)$.
This distribution is used to generate the output token and update the hidden state vector.
To generate the token, we either sample a word from the current state
or copy a word from the input using attention.
Allowing our model to copy from the input is helpful for our task, since many of the entities are repeated verbatim in both input and output.
We refer the reader to~\citet{wang2016recombination} for more details. 


 

\paragraph{Evaluation.}

\begin{table*}
  \begin{center}
  \begin{tabular}{p{0.49\textwidth} p{0.45\textwidth}} 
    \toprule 
    \bf \bf Input formula & \bf \bf Generated perspective \\ \midrule
7 $\times$ the cost of an employee $\times$ a week & 7 times the cost of employing one person for one week \\
1/10 $\times$ the cost of an employee $\times$ the population of California $\times$ the time taken for a football game & one tenth the cost of an employee during a football game by the population of California \\
1 $\times$ coffee consumption $\times$ a minute $\times$ population of the world	& the amount of coffee consumed in one minute on the world \\
6 $\times$ weight of a person $\times$ population of California	& six times the weight of the people who is worth \\
    \bottomrule
  \end{tabular}

  \end{center}
  \caption{\label{tab:examples-generation} Examples of perspectives generated by the sequence-to-sequence RNN.\@
  The model is able to capture rephrasings of fact descriptions and reordering of the facts.
  However, it often confuses prepositions and, very rarely, can produce nonsensical utterances.}
\end{table*}

We split the perspective description dataset into a training and test set such
that no formula in the test set contains the same set of tuples as a formula
in the training set.%
\footnote{Note that formulas with the same set of tuples can occur multiple times in the either the training or test set with different multipliers.}
\reftab{results-generation} compares the performance of the baseline and sequence-to-sequence RNN using BLEU.\@
The sequence-to-sequence RNN performs significantly better than the baseline,
producing more natural rephrasings.
\reftab{examples-generation} shows some output generated by the system (see \reftab{examples-generation}).


\section{Human evaluation}
\label{sec:human-evaluation}

\begin{table*}
  \begin{center}
  \begin{minipage}[c]{0.36\textwidth}
      \centering
      \vfill{}
  \begin{tabular}{lll}
    \toprule
    \textbf{\systemo{}} & \textbf{\systemb{}} & \\
    \multicolumn{2}{c}{\textbf{perspective rated useful?}} & \textbf{\#} \\ \midrule
    Yes & Yes & 31 \\
    Yes & No & 63 \\
    No & Yes & 61 \\
    No & No & 56 \\
    \bottomrule
  \end{tabular}

      \subcaption{\label{tab:results-errors}
      A summary of the number of times the perspective generated by \systemo{} or \systemb{} was rated useful by a majority of crowdworkers.}
      \vspace{2em}
  \begin{tabular}{lll}
    \toprule
    \textbf{Cause of error} & \textbf{\#} \\ \midrule
    Proximity & 9 \\
    Familiarity & 6 \\
    Compatibility & 8 \\
    Similarity & 49 \\ \cmidrule{1-1}
    Generation & 24 \\
    Other & 14 \\
    Invalid mention & 7 \\ \midrule
    \textbf{Total} & 117 \\
    \bottomrule
  \end{tabular}

      \subcaption{\label{tab:analysis-errors} An analysis of errors produced by \systemo{} when its perspectives were not rated useful. 
      Errors caused by poor formula selection are further categorized by selection criteria violated.
      }
      \vfill{}
  \end{minipage}
  \hfill
  \begin{minipage}[c]{0.56\textwidth}
 \begin{tabular}{p{0.10\textwidth} p{0.81\textwidth}} 
    \toprule 
    \bf Cat. & \bf Mention \\ \cmidrule{2-2}
                 & \bf \systemo{} perspective (\textit{vs.} \systemb{}) \\ \midrule
    Prox.
    & \dots ready to ship about \textbf{2,300 miles} across the Pacific to the mainland \dots \\ \cmidrule{2-2}
      & three times the distance from San Francisco to Los Angeles 
      (\textit{vs.} the distance from San Francisco to Dallas TX).
      \\ \midrule
    Sim.
    & China had disposed of about \textbf{100,000 tons} of CFCs'' \dots \\ \cmidrule{2-2}
      & one fifth of the weight of garbage produced in the United States by the population of Texas in one week.
      (\textit{vs.} the average food wasted every year). \\
      \\ \midrule
    Fam.
    & \dots the project could save New England ratepayers \textbf{\$4.6 billion} in energy costs over 25 years. \\ \cmidrule{2-2}
        & one eighth the cost of employing the population of Asia for one hour.
         (\textit{vs.} the construction cost of The Cosmopolitan in Las Vegas.) 
        \\ \midrule
    Comp.
    & Hominids started shaping stone tools about \textbf{2.6 million years} ago. \\ \cmidrule{2-2}
      & 5 times the total time taken to build the number of cars registered.
      (\textit{vs.} 17000 times the average lifetime for a tree).
      \\ 
    \bottomrule
  \end{tabular}

        \subcaption{\label{tab:examples-errors} Examples of errors categorized by the criteria defined in \refsec{selection}.}
  \end{minipage}
      \caption{\label{tab:rate} Results of an end-to-end human evaluation of the output produced by our perspective generation system (\systemo{}) and a baseline ($\systemb$) that picks the numerically closest tuple in the knowledge base for each mention.}
  \end{center}
\end{table*}

\begin{table*}[t]
  \begin{center}

\begin{tabular}{l p{0.45\textwidth} p{0.45\textwidth}}
\toprule 
& \bf Mention & \bf Perspective (that's about\dots) \\ \midrule
+ & In 2007, Turkmenistan exported \textbf{50 billion cubic meters} of gas to Russia. &
the amount of oil produced by the US during a lifetime \\
+ & It can carry up to 10 nuclear warheads and has a range of \textbf{8,000 km}. &
the distance from San Francisco to Beijing \\
- & the \textbf{2.7 million square feet} that Mission Bay's largest developer is entitled to build & twice the area of forest logged in a minute  \\
- & Las Vegas Sands claims the \textbf{10.5 million square feet} is the largest building in Asia. &
one half of an area of an average farm \\
\bottomrule
\end{tabular}

  \end{center}
  \caption{\label{tab:examples-framing}
  Examples of perspectives generated by our system that frame the mentioned quantity to be larger or smaller (top to bottom) than initially the authors thought.}
\end{table*}

In addition to the automatic evaluations for each component of the system, we also ran an end-to-end human evaluation on an independent set of 211 mentions collected using the same methodology described in \refsec{dataset}.
Crowdworkers were asked to choose between perspectives generated by our full system ($\systemo$) and those generated by the baseline of picking the numerically closest tuple in the knowledge base ($\systemb$). 
They could also indicate if either both or none of the shown perspectives appeared useful.\footnote{%
Crowdworkers were paid \$0.06 per to choose a perspective for each mention.
Each mention and set of perspectives were presented to 5 crowdworkers.}

\reftab{rate} summarizes the results of the evaluation and an error analysis conducted by the authors.
Errors were characterized as either being errors in generation (e.g.\ \reftab{examples-generation}) or violations of the criteria in selecting good formulas described in \refsec{selection} (\reftab{examples-errors}).
The \textit{other} category mostly contains cases where the output generated by \systemo{} appears reasonable by the above criteria but was not chosen by a majority of workers.
A few of the mentions shown did not properly describe a numeric quantity, e.g.\ ``\dots claimed responsibility for a \textit{2009 gun} massacre \dots'' and were labeled \textit{invalid mentions}.
The most common error is the selection of a formula that is not contextually relevant to the mentioned text because no such formula exists within the knowledge base (within an order of magnitude of the mentioned value): a larger knowledge base would significantly decrease these errors.
\section{Related work and discussion}
\label{sec:related}




We have proposed a new task of perspective generation.
Compositionality is the key ingredient of our approach,
which allows us synthesize information across multiple sources of information.
At the same time, compositionality also poses problems for both
formula selection and description generation.

On the formula selection side,
we must compose facts that make sense.
For semantic compatibility between the mention and description,
we have relied on simple word vectors \citep{mikolov2013efficient},
but more sophisticated forms of semantic relations on larger units of text
might yield better results \citep{bowman2015large}.

On the description generation side,
there is a long line of work in generating natural language descriptions of
structured data or logical forms
\citet{wong07generation,chen08sportscast,lu12probabilistic,angeli10generation}.
We lean on the recent developments of neural sequence-to-sequence models
\citep{sutskever2014sequence,bahdanau2014neural,luong2015translation}.
Our problem bears some similarity to the semantic parsing work of \citet{wang2015overnight},
who connect generated canonical utterances (representing logical forms) to real utterances.

If we return to our initial goal of helping people understand numbers,
there are two important directions to explore.
First, we have used a small knowledge base, which limits the coverage of perspectives
we can generate.  Using Freebase \citep{bollacker2008freebase} or even open
information extraction \citep{fader11reverb} would dramatically increase the number of facts
and therefore the scope of possible perspectives.

%
Second, 
while we have focused mostly on basic compatibility,
it would be interesting to explore more deeply how the juxtaposition of facts affects framing. 
\reftab{examples-framing} presents several examples generated by our system that frame the mentioned quantities to be larger or smaller than the authors originally thought.
We think perspective generation is an exciting setting to study aspects of numeric framing~\citep{teigen2015framing}.

\paragraph{Reproducibility}
All code, data, and experiments for this
paper are available on the CodaLab platform at
{\small \url{https://worksheets.codalab.org/worksheets/0x243284b4d81d4590b46030cdd3b72633/}}.

\section*{Acknowledgments}
We would like to thank 
Glen Chiacchieri for providing us information about the Dictionary of Numbers,
Maneesh Agarwala for useful discussions and references,
Robin Jia for sharing code for the sequence-to-sequence RNN,
and the anonymous reviewers for their constructive feedback.
This work was partially supported by the Sloan Research fellowship to the second author.

\bibliographystyle{acl_natbib}
\bibliography{refdb/all}

\end{document}